\documentclass[runningheads]{llncs}

\RequirePackage[hyphens]{url} % break urls
\PassOptionsToPackage{hyphens}{url}\usepackage{hyperref} % break urls

\usepackage[T1]{fontenc}
\usepackage{graphicx}
\usepackage{stackengine}
\usepackage{layouts}

\usepackage{array} % stretch/shrink table width
\usepackage{pbox} % line breaks in tables
\usepackage{amsmath} % for \text in math mode
\usepackage{amssymb}
\usepackage[super]{nth}
\usepackage{marvosym} % for email symbol \Letter
\usepackage{capt-of}% or \usepackage{caption}
\usepackage{subcaption} 

\usepackage{booktabs}
\usepackage{tabularray}
\usepackage{multirow}
\usepackage{comment}

\usepackage[table]{xcolor}
\usepackage{pgf}
\usepackage{collcell}
\usepackage{layouts}

\newcommand*{\MinNumber}{40}%
\newcommand*{\MaxNumber}{200}%

\newcommand{\ApplyGradient}[1]{%
  \pgfmathsetmacro{\PercentColor}{100*(#1-\MinNumber)/(\MaxNumber-\MinNumber)}%
  \edef\x{\noexpand\cellcolor{black!\PercentColor}}\x\textcolor{black}{#1}%
}
\newcolumntype{R}{>{\collectcell\ApplyGradient}{c}<{\endcollectcell}}

\usepackage{enumitem}
\setitemize{label=\textbullet}
\usepackage{float}
\usepackage{orcidlink}

\begin{document}

% Requirements after decision: 8.5 Pages + 2 Pages Refs

\title{Label-free estimation of clinically relevant performance metrics under distribution shifts}
\titlerunning{Label-free estimation of clinically relevant performance metrics}

\author{
    Tim~Flühmann\orcidlink{0009-0005-0382-4675}\inst{1,2,3}\textsuperscript{(\Letter)} \and
    Alceu Bissoto\orcidlink{0000-0003-2293-6160}\inst{1,2,3} \and
    Trung-Dung Hoang\orcidlink{0009-0005-4743-3246} \inst{1,2,3} \and
    Lisa~M.~Koch\orcidlink{0000-0003-4377-7074}\inst{1,2,3}\textsuperscript{(\Letter)}
}
\authorrunning{Flühmann et al.}
\institute{
    University of Bern, Switzerland \and
    Department of Diabetes, Endocrinology, Nutritional Medicine and Metabolism, Inselspital, Bern University Hospital, University of Bern, Switzerland \and
    Diabetes Center Berne, Switzerland \\
    \email{\{tim.fluehmann,lisa.koch\}@unibe.ch}
}

\maketitle

\setcounter{footnote}{0}
\begin{abstract}
Performance monitoring is essential for safe clinical deployment of image classification models. However, because ground-truth labels are typically unavailable in the target dataset, direct assessment of real-world model performance is infeasible.
State-of-the-art performance estimation methods address this by leveraging confidence scores to estimate the target accuracy.
Despite being a promising direction, the established methods mainly estimate the model's accuracy and are rarely evaluated in a clinical domain, where strong class imbalances and dataset shifts are common.
Our contributions are twofold: First, we introduce generalisations of existing performance prediction methods that directly estimate the full confusion matrix. Then, we benchmark their performance on chest x-ray data in real-world distribution shifts as well as simulated covariate and prevalence shifts.
The proposed confusion matrix estimation methods reliably predicted clinically relevant counting metrics on medical images under distribution shifts. However, our simulated shift scenarios exposed important failure modes of current performance estimation techniques,
calling for a better understanding of real-world deployment contexts when implementing these performance monitoring techniques for postmarket surveillance of medical AI models.\footnote{Code available at \url{https://github.com/mlm-lab-research/clin_perf_est}}

\end{abstract}

\section{Introduction}

Deep learning for medical image classification exhibits excellent performance in controlled settings \cite{Liu_Faes_Kale_Wagner_Fu_Bruynseels_Mahendiran_Moraes_Shamdas_Kern_et_al._2019}, but distribution shifts in the target domain may cause silent failures in real-world applications \cite{Varoquaux_Cheplygina_2022,oakden2020hidden}. For safe deployment in clinical domains, continuous performance monitoring is crucial. Several methods have been proposed to estimate model classification performance on unlabelled target datasets, enabling clinicians to anticipate model failures before they affect patients. 
Some performance estimation approaches estimate accuracy based on the distance between source and target distribution \cite{guillory2021predicting}, or train a reverse model on the pseudo-labelled test data to evaluate reverse performance on the source distribution \cite{Fan_Davidson_2006}.
Here, we focus on performance estimation based on the model's confidence scores \cite{guillory2021predicting,białek2024estimatingmodelperformancecovariate,garg2021leveraging,li2022estimating,kivimaki2025confidence,kivimaki2025performance}. They have shown the best trade-off between accuracy and computational efficiency \cite{garg2021leveraging,li2022estimating,elsahar2019annotate}, as they require neither retraining, other labelled datasets, nor ensemble agreement \cite{baek2022agreement}. Despite their potential, it remains unclear whether confidence-based performance-estimation methods apply in the clinical setting. Most were proposed and evaluated outside the clinical domain and focus on predicting only accuracy, which is often an inadequate metric for clinical tasks \cite{maier2024metrics}. Instead, validating medical image classification models requires a suite of clinically relevant metrics depending on the domain of interest, such as precision, recall, etc. Currently, no benchmark exists for these metrics, and we are aware of only one first naive approach \cite{kivimaki2025performance} to estimate these metrics without access to labelled test data.

\begin{figure}[!t]
    \centering
    \includegraphics[width=\linewidth]{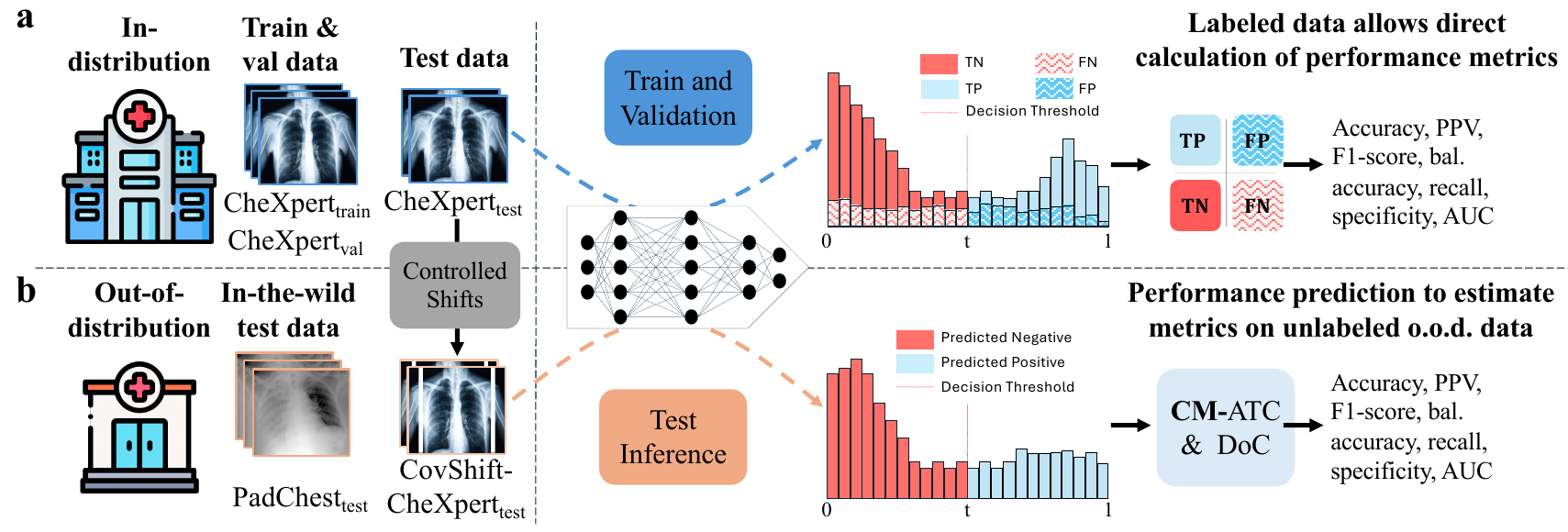}
    \caption{
    Performance monitoring in a clinical setting. (\textbf{a}) An idealised, in-distribution scenario with labelled test data. (\textbf{b}) Our method for estimating metrics in the more realistic, out-of-distribution, unlabelled setting.} 
    \label{fig:graphical_abs}
\end{figure}

In this paper, we propose a generalisation of two popular performance estimation techniques \cite{guillory2021predicting,garg2021leveraging}, which allows us to go beyond accuracy and
reliably estimate the full confusion matrix (i.e., true positives (tp), false positives (fp), true negatives (tn), and false negatives (fn)), and thus any counting metric (e.g., recall, PPV) as well as the multi-threshold area under the ROC curve (AUC).
We then present a comprehensive benchmark for confidence-based performance estimation in chest x-ray data, predicting a range of clinically relevant performance metrics (see overview in Fig.\,\ref{fig:graphical_abs}). We study real-world distribution shifts as well as the effects of covariate and prevalence shifts, which are common in medical data and can impair the clinical translation of medical classification models. 

\section{Background: performance estimation without labels}
We define a binary classification problem with targets $y\in\{0, 1\}$, where models $f(x)$ are trained on inputs $\{(x_i,y_i)\}_{i=1}^m \sim \mathcal{D}_{\mathrm{train}}$ and validated on $\mathcal{D}_{\mathrm{val}}$, which is assumed to be in-distribution (i.d.) with respect to $\mathcal{D}_{\mathrm{train}}$. Predictions $\widehat{y}$ are obtained by applying a decision threshold $t \in [0, 1]$ to the model's sigmoid output $\tilde{s}_i = \sigma(f(x_i))$, which also represents the model's positive class confidence score. To obtain the confidence in the predicted class, we set $s_i=\mathbb{I}_{\{\tilde{s}_i\ge t\}}\, \tilde{s}_i
      + \mathbb{I}_{\{\tilde{s}_i< t\}}\,(1 - \tilde{s}_i)$. 
Our objective is to assess the model’s performance through metrics like accuracy, recall, PPV, and AUC on a previously unseen, potentially out-of-distribution  (o.o.d.) dataset $\{(x_i,y_i)\}_{i=1}^n\sim \mathcal{D}_{\mathrm{test}}$, for which ground-truth labels are not available. Furthermore, we define the sets of positive and negative class predictions for distributions ${d\in\{{\mathrm{val},\mathrm{test}}}\}$ as $I^+_{d}=\{\tilde{s}_{i}\,|\,\tilde{s}_{i} \geq t,\, x_i\sim \mathcal{D}_{d}\}$ and $I^-_{d}=\{1-\tilde{s}_{i}\,|\,\tilde{s}_{i} < t,\, x_i\sim \mathcal{D}_{d}\}$ with cardinalities $n^+_{d}$ and $n^-_{d}$ respectively.

\subsubsection{Confidence Based Performance Estimation (CBPE).} 
A model is considered calibrated when its confidence score reflects the true probability of class 1; formally, $P(Y=1|\tilde{S}=\tilde{s})=\tilde{s}, \forall \,\tilde{s}\in[0, 1]$ \cite{pmlr-v70-guo17a}. For calibrated models, accuracy can thus be estimated by the average confidence \cite{kivimaki2025confidence,hendrycks2017a} $\widehat{\text{acc}}_{\text{CBPE}} = \frac{1}{n}\sum_{i=1}^ns_i.$

This approach has recently been generalised for other counting metrics based on the confusion matrix \cite{kivimaki2025performance}. CBPE builds on the observation that scores with $\tilde{s}_i \geq t$ are either tp or fp, whereas scores with $\tilde{s}_i < t$ are either tn or fn. Averaging the subsets $I_{\text{test}}^+$ and $I_{\text{test}}^-$ provides estimates for positive predictive value (PPV) and negative predictive value (NPV):

\begin{equation}
\widehat{\text{PPV}}_{\text{CBPE}} = \mathbb{E}_{s\sim I_{\text{test}}^+}[s]\,, \,\,\
    \widehat{\text{NPV}}_{\text{CBPE}} = \mathbb{E}_{s\sim I_{\text{test}}^-}[s].
\end{equation}

All counting metrics can then be estimated from the confusion matrix point estimates, which are derived from:
\begin{equation}
\label{eq:cm_estimation}
\begin{split}
    \hat{\text{tp}} &=   n^+_{\text{test}}\cdot \widehat{\text{PPV}}, \,\,\,\,\,\hat{\text{fp}} = n^+_{\text{test}} - \hat{\text{tp}}, \\ 
    \hat{\text{tn}} &=  n^-_{\text{test}}\cdot\widehat{\text{NPV}},\,\,\,\, 
    \hat{\text{fn}} = n^-_{\text{test}} - \hat{\text{tn}}.
\end{split}
\end{equation}

\subsubsection{Average Threshold Confidence (ATC).}
\label{sec:background_atc}
ATC \cite{garg2021leveraging} predicts the model accuracy on the test distribution $\mathcal{D}_{\mathrm{test}}$ by computing the proportion of samples with scores $s$ exceeding a learned threshold $t_{\text{ATC}}$. This threshold is determined on the validation set $\mathcal{D}_{\mathrm{val}}$ such that the proportion of scores above $t_{\text{ATC}}$ matches the empirical accuracy on $\mathcal{D}_{\mathrm{val}}$: $ \mathbb{E}_{x \sim \mathcal{D}_{\mathrm{val}}}[\mathbb{I}[s > t_{\text{ATC}}]] = \mathbb{E}_{(x, y) \sim \mathcal{D}_{\mathrm{val}}}[\mathbb{I}[\widehat{y} = y]].$

The accuracy on $\mathcal{D}_{\mathrm{test}}$ is then estimated as the fraction of test samples with confidence scores above the learned threshold $t_{\text{ATC}}$.

\subsubsection{Difference of Confidences (DoC).}
\label{sec:background_doc}
DoC \cite{guillory2021predicting} can be used as a confidence-based performance estimation method that estimates test accuracy via:

\begin{equation}
    \widehat{\text{acc}}_\text{DoC} =  \text{acc}_{\text{val}}- \Delta ; \quad \Delta = \mathbb{E}_{x\sim \mathcal{D}_{\mathrm{val}}}[s] -  \mathbb{E}_{x\sim \mathcal{D}_{\mathrm{test}}}[s]
\end{equation}

where $\text{acc}_{\text{val}}$ is the observed accuracy in $\mathcal{D}_{\mathrm{val}}$. In other words, the scores on the test distribution are re-calibrated by the amount that the average validation confidence deviates from the validation accuracy. 

\section{Performance estimation beyond accuracy}

So far, estimating clinically relevant metrics beyond accuracy was only possible with the naive confidence-based CBPE. Here, we propose a method to extend ATC and DoC to estimate PPV (precision) and NPV. From these estimates, we can compute individual entries of the confusion matrix through Eq. \eqref{eq:cm_estimation}, allowing us to then predict any counting metric from the confusion matrix estimators; for example, recall is estimated as $\hat{\text{tp}}/(\hat{\text{tp}}+\hat{\text{fn}})$. To approximate the AUC, we evaluate the true-positive rate and false-positive rate at 100 decision thresholds based on score quantiles and then numerically integrate the resulting ROC curve.

\subsubsection{Confusion Matrix Estimation via Average Threshold Confidence (CM-ATC).}

To extend the ATC framework toward estimating elements of the confusion matrix, we build on an idea from a previously proposed variant of ATC~\cite{li2022estimating}.  
They applied separate thresholds $t_{\text{ATC}}^{+}$ to positive and $t_{\text{ATC}}^{-}$ to negative predicted cases, learned on $I_{\text{val}}^{+}$ and $I_{\text{val}}^{-}$, respectively. 
In~\cite{li2022estimating}, the positive and negative prediction sets were never used individually, but instead, accuracy was estimated by aggregating and counting all samples that exceeded their respective thresholds.

Here, we propose to use the positive $I_{\text{test}}^{+}$ and negative $I_{\text{test}}^{-}$ prediction sets in isolation along with their respective learned thresholds. The two class-specific thresholds are calculated such that the fraction of scores above $t_{\text{ATC}}^{+}$ on $I_{\text{val}}^{+}$ equals the validation PPV, and, analogously, the fraction of scores above $t_{\text{ATC}}^{-}$ on $I_{\text{val}}^{-}$ equals the validation NPV. We then define the following estimators for PPV and NPV, similarly to the original accuracy estimator described in Sec.\,\ref{sec:background_atc}:
\begin{equation}
\begin{split}
    \widehat{\text{PPV}}_{\text{CM-ATC}} &= \mathbb{E}_{s\sim I_{\text{test}}^{+}}[\mathbb{I}[s>t_{\text{ATC}}^+]]\,,\,\,\,\,\widehat{\text{NPV}}_{\text{CM-ATC}}=\mathbb{E}_{s\sim I_{\text{test}}^{-}}[\mathbb{I}[s>t_{\text{ATC}}^-]].
\end{split}
\end{equation}
With estimates for PPV and NPV, we analogously estimate the confusion matrix through Eq. \eqref{eq:cm_estimation} to further estimate any counting metric of interest.

\subsubsection{Confusion Matrix Estimation via Difference of Confidences (CM-DoC)} 

Similarly to the ATC variant, \cite{li2022estimating} have extended DoC to re-calibrate the scores in $I_{\text{test}}^{+}$ and $I_{\text{test}}^{-}$ separately before estimating test accuracy. On $I_{\text{test}}^{+}$, the scores are offset by the gap between validation PPV and the mean confidence on $I_{\text{val}}^{+}$. Analogously, on $I_{\text{test}}^{-}$ by the gap between validation NPV and the mean confidence on $I_{\text{val}}^{-}$.

Following the approach of CM-ATC, we now also extend the DoC method to estimate the confusion matrix elements. After calculating the offsets $\Delta^c := \mathbb{E}_{s \sim I_{\text{val}}^c}[s] - \mathbb{E}_{s \sim I_{\text{test}}^c}[s]$ for \( c \in \{+, -\} \) and the realized $\text{PPV}_{\text{val}}$ and $\text{NPV}_{\text{val}}$ on validation, we can get estimates on the test set through:

\begin{equation}
    \widehat{\text{PPV}}_{\text{CM-DoC}} =  \text{PPV}_{\text{val}} -\Delta^+;\,\,\,\,\,
        \widehat{\text{NPV}}_{\text{CM-DoC}} = \text{NPV}_{\text{val}} -\Delta^-.
\end{equation}
From here, we again get the confusion matrix estimates from Eq. \eqref{eq:cm_estimation} and calculate estimates for the metrics.

\section{Benchmark setup}
With the methods for estimating performance metrics in place, we now set up their comparison on real-world and controlled distribution shifts on medical images. We compare CBPE and our proposed CM-ATC and CM-DoC. In addition, we include a naive baseline for ATC and DoC: we take the original ATC and DoC formulations and substitute accuracy with the metric of interest. 
In preliminary experiments, we have also analysed the impact of calibration (using temperature scaling \cite{pmlr-v70-guo17a} and class-wise temperature scaling \cite{li2022estimating}). As we had found no conclusive improvements (see Supplementary Fig. \,S1), we excluded an analysis of calibration techniques from the scope of this paper.
We estimate a wide variety of metrics, covering prevalence-dependent (accuracy, PPV, F1-score), prevalence-independent counting metrics (balanced accuracy, recall, specificity), and a multi-threshold metric (AUC). 
Furthermore, we monitor the model's calibration using the Root Brier Score (RBS) and Adaptive Calibration Error (ACE).

\subsection{Chest x-ray distribution shifts in the wild}

First, we benchmark the performance estimators on real-world distribution shifts in chest x-ray data using three publicly available datasets from different cohorts: \textbf{CheXpertPlus} \cite{chambon2024chexpert}, \textbf{PadChest} \cite{bustos2020padchest}, and \textbf{ChestX-Ray8} (NIH) \cite{wang2017chestx}. They consist of 223,228, 160,861, and 112,120 chest radiographs, respectively. 
For each dataset, we set aside roughly 22,000 images for validation and testing (CheXpertPlus: 90/10/10; PadChest and NIH: 60/20/20) and use the remainder to train separate binary classifiers for three target conditions: Pleural Effusion, Cardiomegaly, and Pneumothorax.
For each model, we estimate the performance metrics on the held-out i.d. test set, as well as on the two o.o.d. test sets. 
To evaluate the generalization capabilities of the performance estimation methods, we compute the difference between the estimated and realized performance metrics and report the mean absolute error (MAE). 

\subsection{Controlled distribution shifts in chest x-rays}
\label{Section:controlled_label_shift}
Next, we investigate the behaviour of performance prediction methods under controlled distribution shifts. For this, we focus on the task of detecting Pleural Effusion in the CheXpertPlus dataset and simulate covariate shifts and prevalence shifts in the held-out test set. 

To introduce \textbf{covariate shift}, we artificially modify the images by adding a visual artefact (two lateral vertical white bars) that is positively correlated with the positive class \cite{sun2023right}. The model is then trained on this modified data, allowing it to learn a spurious correlation between the artefact and the target label. As a consequence, the model produces high-confidence predictions for the majority groups (i.e., label 1 with artefact present and label 0 without artefact), while generating lower-confidence predictions for the minority groups. 
For the training and validation sets, we set the proportion of majority samples to 80\%, encouraging the model to rely on the spurious feature. Test sets are constructed by sampling 1000 images with varying proportions of majority samples from $0\%$ to $100\%$ and consequently minority samples from $100\%$ to $0\%$. We keep the class distribution consistent between the validation and test sets to prevent additional prevalence shifts from confounding the analysis.  We repeat the test set construction 50 times for each shift strength to reduce sampling variability, and the estimations are averaged over all repetitions.

To simulate \textbf{prevalence shift}, we repeatedly sample 1000 instances from the test set while targeting a positive class prevalence ranging from $5\%$ to $95\%$. Since the overall prevalence of Pleural Effusion in the CheXpertPlus dataset is relatively high, at 38\%, we can simulate label shifts in both directions, toward lower and higher prevalence levels without restricting the dataset size too much. We perform 50 resampling iterations at each prevalence level and compute averaged realised and estimated performance metrics~\cite{kivimaki2025confidence}. 

\section{Results}
\subsection{CM estimation methods perform best in the wild}
\label{sec:wild-results}

\begin{figure}[t]
    \centering
    \includegraphics[width=0.85\linewidth]{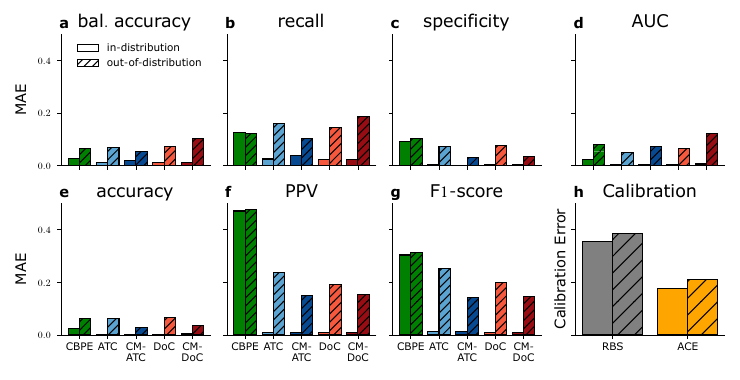}
    \caption{(\textbf{a}-\textbf{g}) MAE evaluated on i.d. and o.o.d. data represented by solid and lined bars, respectively. (\textbf{h}) shows the mean RBS and ACE over all models. On average, CM methods outperformed the other estimators. Exact numerical values are reported in the Supplementary Table S2.}
    \label{fig:MAE_wilds}
\end{figure}

The trained models performed comparably to the state-of-the-art in terms of classification performance \cite{cohen2020limits}, see Supplementary Table S1 for details. 
Overall, across both i.d. and o.o.d. scenarios and the different metrics in Fig.\,\ref{fig:MAE_wilds}, our CM estimators outperformed the other methods, with CM-ATC performing best.
Several prior studies, including CBPE, ATC, and DoC, have focused on accuracy (Fig.\,\ref{fig:MAE_wilds}\,\textbf{e}). Here, all methods worked well in-distribution (mean MAE of $(0.7\pm0.7)\cdot10^{-2}$). Estimating accuracy in o.o.d. datasets was more difficult for all methods with a mean MAE of $0.05~\pm~0.02$.

To estimate metrics beyond accuracy, only CBPE has been proposed before. While estimation of balanced accuracy, specificity, and AUC performed comparably to accuracy, performance dropped considerably for estimating recall, and failed dramatically for PPV and F1-score. CBPE's estimation error could be attributed to the significant calibration error (Fig.\,\ref{fig:MAE_wilds}\,\textbf{h}). In contrast, the ATC and DoC approaches led to consistently very low estimation error for all metrics in i.d. settings. O.o.d. performance varied, with CM-ATC generalising best overall. Since we have not explicitly quantified the distribution shift between the chest X-ray datasets, the observed drop is not straightforward to interpret.

\subsection{Performance estimation methods can capture the impact of covariate shift}

\begin{figure}[t]
    \centering
    \includegraphics[width=0.9\linewidth]{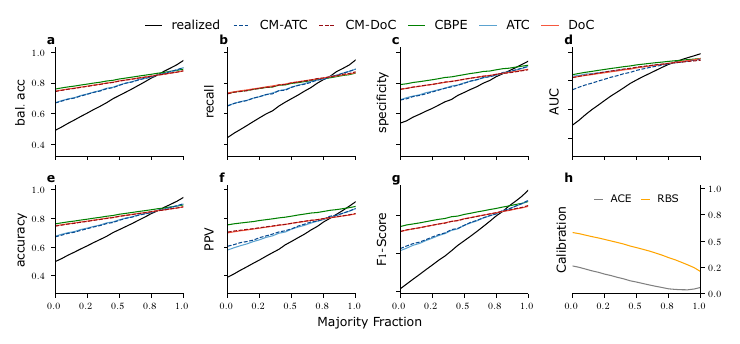}
    \caption{(\textbf{a-g}) The estimated performance metrics capture the decline in performance under the simulated covariate shift, yet show overconfident performance towards minority groups in the test set. (\textbf{h}) Calibration error rises significantly for the minority group.}
    \label{fig:controlled_covariate_pleff}
\end{figure}

Next, we introduced covariate shifts using synthetic artefacts (see Sec.\,\ref{Section:controlled_label_shift}).
By design, actual model performance deteriorated when increasing the proportion of minority groups (black curves in Fig.\,\ref{fig:controlled_covariate_pleff}\,\textbf{a-g}, original majority fraction $p=0.8$). The metric estimators successfully captured this decline, yet still overestimated performance in these cases, and underestimated performance when the majority group was more prevalent than in the original distribution. Once again, (CM-)ATC methods performed best overall, while (CM-)DoC and CBPE likely suffered from model miscalibration on the minority samples (Fig.\,\ref{fig:controlled_covariate_pleff}\,\textbf{h}). The close agreement between the CM estimators and their respective naive counterparts can be attributed to this covariate shift affecting both negative and positive predictions similarly. As a result, the benefit of treating the two groups separately is limited.

\subsection{Metric estimation degrades under prevalence shifts}

\begin{figure}[t]
\centering
\includegraphics[width=0.85\linewidth]{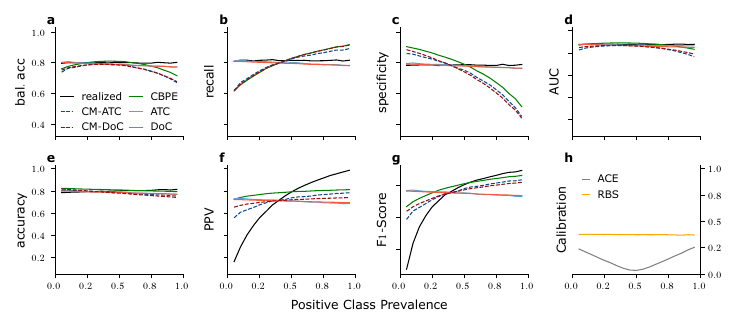}
\caption{(\textbf{a–d}) Prevalence-independent metrics remain constant under label shift; naive estimators perform best, whereas confusion-based ones remain prevalence-sensitive. (\textbf{e–g}) Prevalence-dependent metrics vary under label shifts; confusion-matrix-based methods capture these changes but still show high errors.   (\textbf{h}) Prevalence shift directly affects model calibration.}
\label{fig:controlled_label_shift}
\end{figure}

All estimation methods struggled under prevalence shifts (Fig.\,\ref{fig:controlled_label_shift}), especially for prevalence-dependent metrics (Fig.\,\ref{fig:controlled_label_shift}\,\textbf{e-g}).
As CBPE is highly dependent on model calibration, it only accurately estimated performance when calibration error was low (Fig.\,\ref{fig:controlled_label_shift}\,\textbf{h}). 
The ATC and DoC methods performed best when no shift was present (38\% original prevalence, see Sec.\,\ref{Section:controlled_label_shift}). There, realised (black lines) and estimated performance (coloured) was very close, in line with the i.d. results in Sec.\,\ref{sec:wild-results}.
When the prevalence shifted, the estimates diverged.
The naive ATC and DoC implementations performed well on prevalence-independent metrics, but because they do not rely on class-specific calibration, they could not estimate prevalence-dependent metrics well (mainly visible in Fig.\,\ref{fig:controlled_label_shift}\,\textbf{f,g}).
In contrast, estimators derived from confusion matrix entries (CBPE, CM-ATC and CM-DoC)  performed best for prevalence-dependent metrics, while they perform worse on the prevalence-independent metrics.

\section{Discussion}
In this paper, we proposed methods to estimate a wide range of classifier performance metrics without access to labelled test data. Our proposed estimators outperformed the only existing baseline (CBPE) for monitoring model performance in real-world distribution shifts and a simulated covariate shift. However, simulated prevalence shifts exposed systematic failures of all performance estimation techniques.

Our techniques for label-free performance estimation could be easily implemented in postmarket surveillance frameworks, as they can monitor deployed medical AI algorithms with clinically relevant performance estimators and little computational overhead. However, as we have demonstrated, the accuracy of our estimators depends on the nature of the encountered distribution shifts. Therefore, we recommend that performance monitoring should be accompanied by distribution shift detection \cite{koch2024postmarket,roschewitz2025automaticdatasetshiftidentification}, identification \cite{roschewitz2025automaticdatasetshiftidentification}, and mitigation~\cite{alexandari2020maximum}.  Especially under prevalence shifts, domain adaptation techniques could help counter the systematic negative impact on model calibration.

\begin{credits}
\subsubsection{\ackname} 
This project was supported by the Diabetes Center Berne and strategic funding of the medical faculty of the University of Bern. Calculations were performed on UBELIX, the HPC cluster at the University of Bern.

\subsubsection{\discintname}
The authors have no competing interests to declare that are relevant to the content of this article.
\end{credits}

\bibliographystyle{splncs04_without_urls}
\bibliography{references.bib}

\appendix
\setcounter{figure}{0}
\renewcommand{\thefigure}{S\arabic{figure}}
\setcounter{table}{0}
\renewcommand{\thetable}{S\arabic{table}}

\newpage
\title{Supplementary Material}
\author{}
\institute{}
\authorrunning{Flühmann et al.}
\titlerunning{Label-free estimation of clinically relevant performance metrics}
\maketitle

\section{Calibration}
\label{sec:a2_calib}
\begin{figure}[hb]
    \centering
    \includegraphics[width=0.9\linewidth]{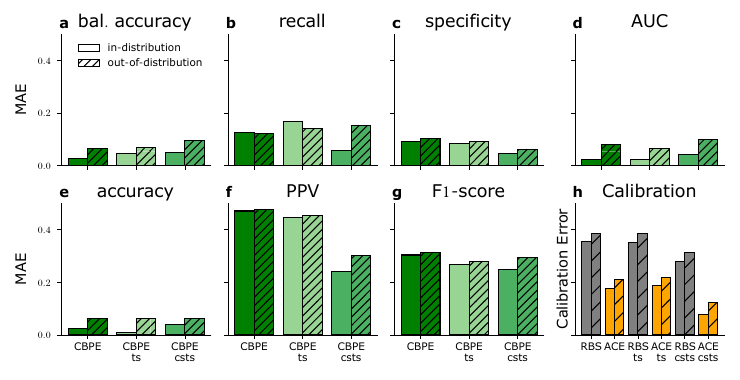}
    \caption{(\textbf{a-g}) CBPE performance estimates, combined with temperature scaling (ts) or class-specific temperature scaling (csts), show no systematic improvement. (\textbf{h}) RBS and ACE quantify calibration quality: ts leaves calibration essentially unchanged, whereas csts substantially reduces both errors.}
    \label{suppfig:calibration}
\end{figure}

\section{Model Performance}
\label{sec:a1_model}
\begin{table}[ht]
\centering
\caption{AUC scores of binary ResNet-50 architectures, trained on different cohorts to classify three pathologies.}
\begin{tabular}{lccc}
\toprule
 & \textbf{ChestXpert} & \textbf{PadChest} & \textbf{NIH} \\
\midrule
Cardiomegaly      & 0.846 & 0.910 & 0.878 \\
Pleural Effusion  & 0.866 & 0.941 & 0.848 \\
Pneumothorax      & 0.854 & 0.850 & 0.851 \\
\bottomrule
\end{tabular}
\label{supp_model_perf}
\end{table}

\newpage
\section{Performance Estimation - Wild shifts}

\begin{table}[ht]
\centering
% \scriptsize
\caption{MAE values of i.d. and o.o.d. performance estimation in the wild. The bold value indicates the lowest MAE in each column.}
\begin{tabular}{l|ccccccc}
\toprule
\textbf{Method} & \textbf{Bal. Acc.} & \textbf{Recall} & \textbf{Specificity} & \textbf{AUC} & \textbf{Accuracy} & \textbf{Precision} & \textbf{F1-score} \\
\midrule
\multicolumn{8}{c}{\textit{in distribution}} \\
\midrule
CBPE    & 0.0289 & 0.1270 & 0.0933 & 0.0242 & 0.0239 & 0.4708 & 0.3036 \\
ATC     & 0.0129 & 0.0252 & 0.0026 & 0.0050 & 0.0031 & 0.0084 & 0.0115 \\
CM-ATC  & 0.0211 & 0.0400 & \textbf{0.0021} & 0.0052 & 0.0035 & 0.0088 & 0.0131 \\
DoC     & \textbf{0.0118} & 0.0240 & 0.0031 & \textbf{0.0044} & \textbf{0.0029} & \textbf{0.0075} & \textbf{0.0100} \\
CM-DoC  & 0.0128 & \textbf{0.0239} & 0.0028 & 0.0063 & 0.0038 & 0.0077 & 0.0106 \\
\midrule
\multicolumn{8}{c}{\textit{out of distribution}} \\
\midrule
CBPE    & 0.0656 & 0.1239 & 0.1029 & 0.0789 & 0.0642 & 0.4761 & 0.3141 \\
ATC     & 0.0688 & 0.1600 & 0.0725 & \textbf{0.0516} & 0.0639 & 0.2363 & 0.2539 \\
CM-ATC  & \textbf{0.0539} & \textbf{0.1047} & \textbf{0.0310} & 0.0736 & \textbf{0.0277} & \textbf{0.1482} & \textbf{0.1404} \\
DoC     & 0.0727 & 0.1463 & 0.0750 & 0.0635 & 0.0677 & 0.1920 & 0.2008 \\
CM-DoC  & 0.1019 & 0.1883 & 0.0347 & 0.1221 & 0.0355 & 0.1551 & 0.1459 \\
\bottomrule
\end{tabular}
\label{supp_performance_wild_shifts}
\end{table}

\end{document}